# 1 Title

Fabric Pneumatic Artificial Muscles Based on the Drawstring Principle

# 2 Authors:

Chendong Liu,[1] Dapeng Yang,[1]* Yiming Dai,[1] Li Jiang,[1] Hong Liu[1]
Affiliations:
[1]State Key Laboratory of Robotics and Systems, Harbin Institute of Technology, Harbin 150001, China.
*Corresponding author. Email: yangdapeng@hit.edu.cn

# 3 Abstract

Pneumatic artificial muscles have wide applications in robotics and industrial fields. Conventional pneumatic artificial muscles generate extra radial deformation during axial contraction, which severely wastes available working space. Inspired by the widely adopted drawstring principle in textile products, this paper proposes a novel drawstring fabric pneumatic artificial muscle (DPAM). Unlike traditional counterparts, the proposed DPAM produces no extra radial deformation during contraction, greatly improving structural compactness. The DPAM exhibits outstanding mechanical performance: a load capacity over 800 times its self-weight, a maximum contraction ratio of 44%, and a power density up to 4.98 kW/kg, alongside excellent scalability. Two representative application scenarios, bionic robots and industrial production lines, are demonstrated to validate its practicability. The DPAM can be easily expanded within a two-dimensional plane, as verified by the fabricated DPAM matrix. This work not only presents a high-performance novel pneumatic artificial muscle but also inspires researchers to draw design inspiration from conventional textile structures to address existing challenges in soft robotics.

# 4 Keywords

# 5 INTRODUCTION

In recent years, as robotic technologies advance toward adaptation to complex environments and safe human–robot interaction, soft robotics has become a prominent research hotspot (*1–6*). Distinct from rigid robots that rely on joints and linkages for locomotion, soft robots are constructed from low-modulus materials to form continuous compliant structures, granting them superior flexibility and environmental adaptability. Such systems demonstrate unique advantages in wearable assistance (*7–10*), minimally invasive surgery (*11–13*), flexible grasping (*14–16*), and locomotion within confined complex environments (*17–19*). Through synergistic design of materials and structures, soft robots can realize multi-degree-of-freedom deformation and sophisticated motion patterns while ensuring human safety, making them a core development direction for next-generation intelligent robotic systems (*20–23*).

As the core actuation units of soft robots, soft actuators directly determine the output capability and kinematic performance of the overall system (*24–26*). Common soft actuation strategies include pneumatic actuation (*27–29*), hydraulic actuation (*30–32*), shape memory material actuation (*33–35*), dielectric elastomer actuation (*17*, *36*, *37*), and magnetically responsive actuation (*38–40*). Among these, pneumatic actuation is widely deployed in soft robotic systems due to its simplicity, low cost, portability, and inherent safety (*41–43*). With elaborately engineered structures, pneumatic soft actuators can achieve bending (*44*, *45*), elongation (*46*, *47*), torsion (*29*, *48*), or contraction motions (*49–52*).

As a typical contractile soft actuator, pneumatic artificial muscles (PAMs) attract extensive attention for their high mechanical output, favorable compliance, and muscle-like force-displacement characteristics (*53–56*). To date, researchers have developed diverse PAM configurations, including McKibben PAMs (*57*, *58*), pleated PAMs (*59*, *60*), flat PAMs (*61*, *62*), pouch motors (*63*, *64*), fabric-lattice PAMs (*65*), X-crossing PAMs (*66*), GRACE PAMs (*67*), and Cavatappi PAMs (*68*, *69*). Nevertheless, nearly all existing PAMs suffer from prominent radial deformation during axial contraction. This radial deformation occupies extra working space and reduces the compactness and integration of the overall mechanical system. When multiple PAMs operate in parallel, mutual structural interference induced by radial deformation complicates assembly and restricts their deployment in wearables, bionic limbs, and other space-limited scenarios. Commercial McKibben PAMs manufactured by FESTO nearly double their radial dimensions upon inflation (*70*), while pleated PAMs exhibit even larger radial expansion ratios (*59*, *60*). For many non-circular PAMs (flat PAMs (*61*, *62*), pouch motors (*63*, *64*), fabric-lattice PAMs (*65*), X-crossing PAMs (*66*), et al), radial and axial deformations are strongly coupled: larger axial stroke corresponds to more severe radial expansion. Although Cavatappi PAMs feature small radial deformation, they require extremely high driving pressure and suffer from insufficient mechanical output performance (*68*, *69*). Therefore, suppressing radial deformation while maintaining high mechanical performance has become a key challenge in the development of pneumatic artificial muscles.

Bionic robotic researchers frequently derive design inspiration from natural organisms. Everyday flexible textile products such as garments, home textiles, and straps can similarly provide innovative ideas for soft robots fabricated from textile materials. Drawing on the drawstring principle used in textiles, this work designs a distinctive drawstring fabric pneumatic artificial muscle (DPAM). By introducing a drawstring-based force transmission mechanism, the volume variation of the airbag under pressurization is converted into axial contraction of drawstrings, eliminating extra radial deformation throughout operation. Compared with conventional designs, the DPAM maintains high force-to-weight ratio (834.7) and large contraction ratio (44%) while drastically enhancing structural compactness and space utilization efficiency. The DPAM is particularly suitable for space-constrained applications such as bionic robots and industrial production lines, and multiple DPAMs can be easily integrated into actuator matrices for large-scale parallel deployment.

# 6 RESULTS

## 6.1 Design Concept and Working Principle

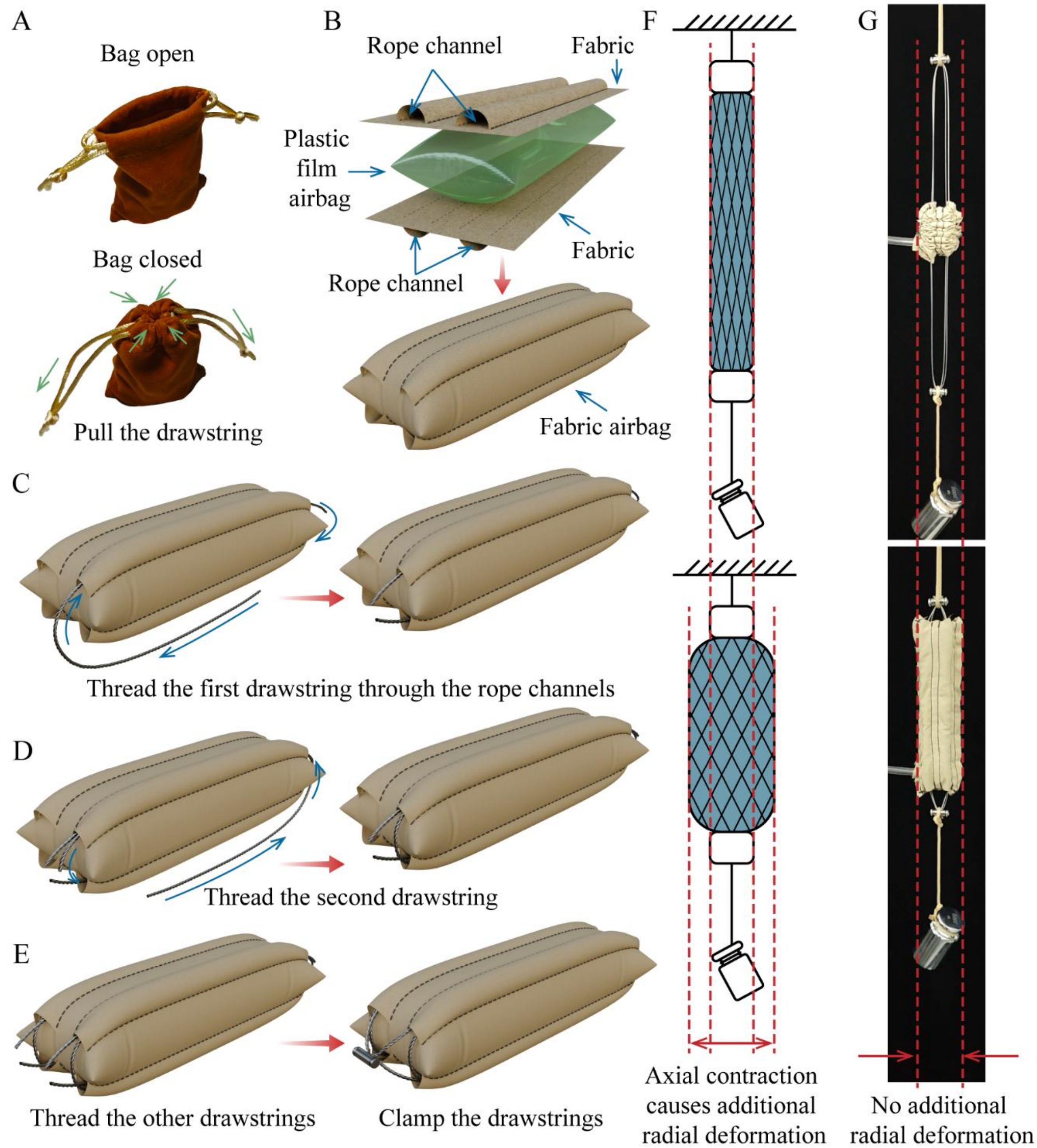

Fig. 1 Drawstring fabric pneumatic artificial muscle (DPAM). (A) Drawstring principle widely adopted in textiles including pockets, cuffs, hood openings, and waistbands. Taking drawstring bags as an example: the bag opens when drawstrings are shortened, and the bag closes when both ends of the drawstrings are pulled apart. (B) Fabrication workflow of the DPAM based on the drawstring principle. Two fabric sheets are cut, each pre-sewn with two semicircular rope channels. A plastic film airbag is formed via thermal sealing,

sandwiched between the two fabric layers, and stitched to encapsulate the airtight airbag into a fabric airbag. (C) The first drawstring is threaded through the rope channels following the illustrated path. (D) The second drawstring is threaded in the opposite direction to the first. (E) The third and fourth drawstrings are threaded identically, with rope terminals fixed by clamping fixtures. (F) Conventional PAMs inevitably generate extra radial expansion during axial contraction, resulting in wasted workspace. (G) Benefiting from its novel structural and actuation mechanism, the DPAM fundamentally eliminates extra radial deformation. When the drawstring ends are stretched, the fabric airbag is constricted, corresponding to the extended state of the DPAM. When compressed air fills the fabric airbag, both ends of the airbag elongate, shortening the drawstring ends and lifting external loads.

With the advancement of soft robotics, textile materials gain increasing popularity for their compliance and low manufacturing cost. Fabric-based soft robots share structural similarities with daily textile goods such as clothing and belts, which serve as a valuable source of design inspiration—analogous to how bionic robots draw ideas from biological organisms. This insight motivates the development of drawstring pneumatic artificial muscles (DPAMs).

The DPAM operates on the drawstring principle pervasive in pocket edges, sleeve cuffs, hoods, and trouser waistbands. Let's take the drawstring bag as an example to illustrate. The drawstring bag shown in Fig. 1A consists of a fabric pocket paired with two drawstrings routed along its perimeter. All we need to do is to pull the drawstring tight, and the pocket will be sealed. Reversing this logic: if an external force (compressed air) inflates and expands the pocket, the drawstring terminals retract, which matches the core contraction function required of pneumatic artificial muscles.

Following this concept, the DPAM prototype is fabricated with two primary materials: outer fabric (khaki) and inner airtight plastic film (green) (Fig. 1B). The plastic film is thermally sealed to form a fully enclosed airbag. The fabric layer is stitched to create four arched rope channels for routing drawstrings. The two fabric layers are sewn together to fully encapsulate the plastic airbag, forming a composite fabric airbag. Fabrics possess high tensile strength but are air-permeable. Plastic films are impermeable but mechanically weak under tension. The composite structure achieves both airtight sealing and high pressure resistance. Four flexible plastic-coated steel wires are utilized as drawstrings. The choice of drawstring material is critical because it serves not only as a load-bearing component but also as a constraint mechanism to regulate and guide the expansion of the fabric airbag, thereby achieving more uniform deformation. Thin plastic-coated steel wires deliver exceptional tensile strength at small diameters, and their smooth coating minimizes frictional loss between ropes and fabric channels. The threading sequence is displayed in Fig. 1C-E. The first drawstring passes through the channels along the path in Fig. 1C, followed by the second drawstring routed oppositely through identical channels (Fig. 1D). The third and fourth drawstrings are threaded following the same procedure, with all drawstring ends secured by clamping fixtures (Fig. 1E).

The DPAM features a simple yet highly effective actuation principle: injecting compressed air into the fabric airbag shortens the drawstring terminals. This fundamentally distinct mechanism eliminates undesirable radial deformation, which is a key advantage that traditional PAMs do not possess. As shown in Fig. 1F, conventional PAMs convert radial expansion into axial contraction via structural coupling, which inherently generates radial deformation. Designers must reserve extra clearance to accommodate this useless radial deformation to avoid mechanical interference, severely limiting system compactness. The DPAM overcomes this limitation fundamentally (Fig. 1G). In the depressurized state, taut drawstrings constrict the airbag into stacked folded layers. Upon pressurization, folded airbag layers inflate and stretch, lengthening the overall airbag and retracting the drawstring ends to produce axial contraction. No radial-to-axial deformation coupling exists in this design, completely eliminating parasitic radial expansion. Supplementary Movie S1 demonstrates the DPAM working principle in full detail. Material specifications are documented in Note S1, and complete fabrication workflows are provided in Note S2, Figure S1, Figure S2, and Movie S2.

## 6.2 Mathematical Modeling and Static Mechanical Testing

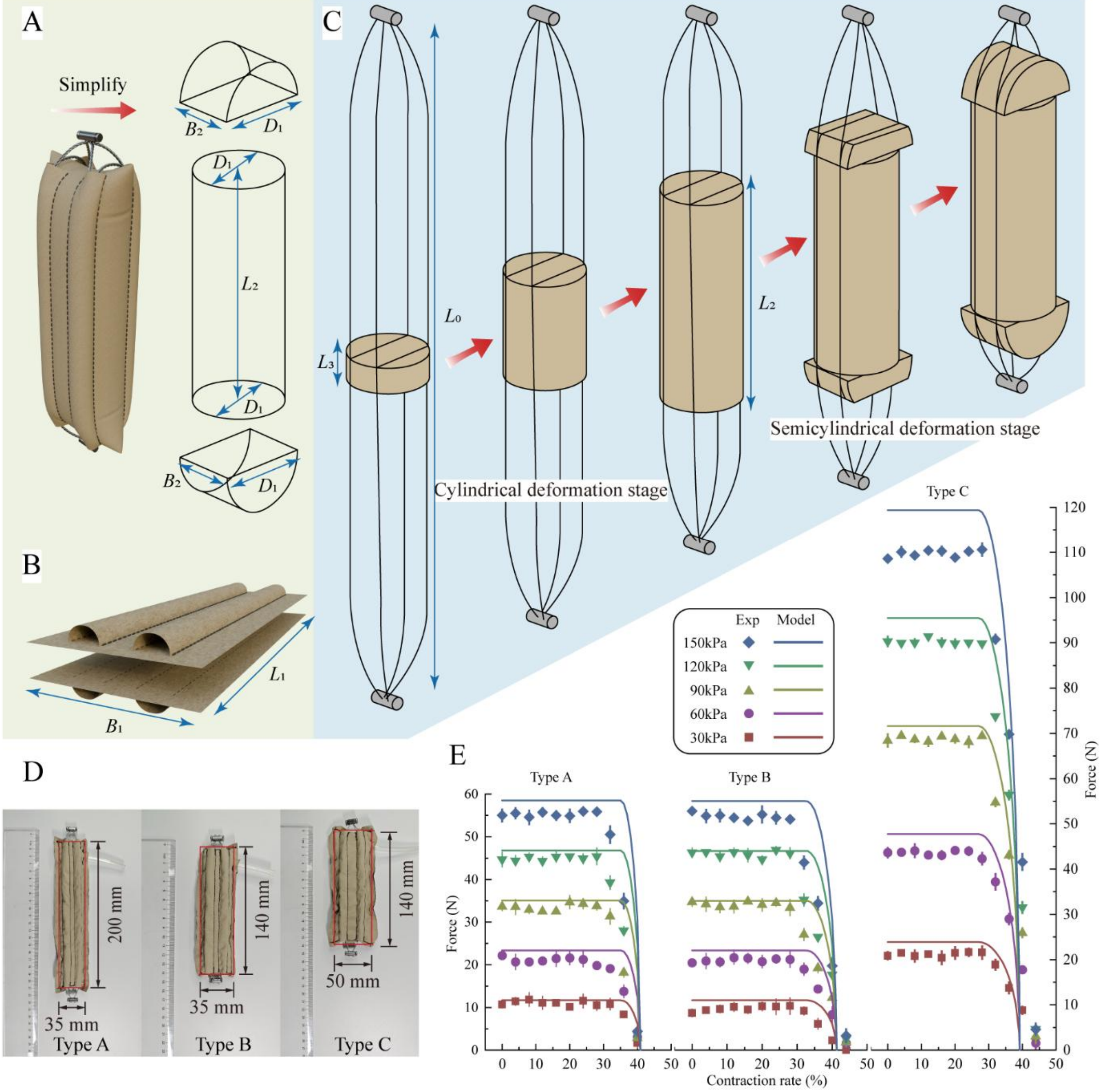

Fig. 2 DPAM mathematical modeling and static mechanical testing. (A) Geometric simplification of the fully inflated DPAM as a central cylinder flanked by two semicylinders. (B) Length and width dimensions of the flat fabric used for airbag fabrication. (C) Simplified two-stage deformation process of the DPAM. In the initial cylindrical stage, the airbag maintains a cylindrical profile as inflation proceeds. As airbag length further increases, semicylindrical end caps gradually form until full expansion is achieved. (D) Three DPAM prototypes with distinct sizes: type A (200 mm length, 35 mm width), type B (140 mm length, 35 mm width), type C (140 mm length, 50 mm width). Type A and B share identical width but differ in length; Type B and C share identical length but differ in width. (E) Output force versus contraction ratio under varying pneumatic pressures for the three DPAM variants. Solid curves denote model predictions, and scattered points represent experimental measurements.

A mathematical model was established to predict the output force of DPAMs with varying geometric dimensions under arbitrary contraction ratios and supply pressures, guiding structural optimization. Due to the irregular inflated geometry of the DPAM airbag, geometric simplification was adopted. The fully inflated airbag is decomposed into a central cylindrical segment and two semicylindrical end caps with uniform diameter $D_1$. The cylinder length is defined as $L_2$, and the semicylinder axial length is $B_2$ (Fig. 2A). Original flat fabric dimensions are denoted as unfolded length $L_1$ and width $B_1$ (Fig. 2B).

Based on observed physical deformation, the actuation cycle is divided into two sequential stages: cylindrical deformation and semicylindrical deformation (Fig. 2C). At the maximum extended DPAM length $L_0$, the fabric airbag reaches its minimum length $L_3$. As compressed air is injected, the airbag elongates gradually to $L_2$ while retaining a cylindrical profile (cylindrical deformation stage). Once airbag length exceeds $L_2$, semicylindrical end caps develop progressively until full airbag expansion (semicylindrical deformation stage). After completing both stages, the DPAM reaches its shortest contracted length. Combining this simplified deformation analysis with the principle of virtual work for airbag expansion, analytical expressions for DPAM output force $F$ are derived separately for the cylindrical and semicylindrical stages.

The calculation formula for the output force during the cylindrical deformation stage is:

$$F = \frac{p}{\pi} B_1^2 \tag{1}$$

The calculation formula for the output force during the semicylindrical deformation stage is:

$$F = pB_1 \sqrt{\frac{B_1^2}{\pi^2} - \left(\frac{\varepsilon L_0 + L_3 + B_1 - L_1}{2}\right)^2} \tag{2}$$

Full model derivation is provided in Supplementary Note S3 and Fig S3.

Three DPAM variants with distinct dimensions are fabricated as illustrated in Fig. 2D: type A ($L_1$=200 mm, $B_1$=35 mm), type B ($L_1$=140 mm, $B_1$=35 mm), type C ($L_1$=140 mm, $B_1$=50 mm). The width of type A is the same as that of type B, but their lengths are different. The lengths of type B and type C are the same, but their widths are different. Comparative testing across these prototypes quantifies the influence of geometric parameters on actuation performance.

Static mechanical characterization is performed under varying supply pressures, with detailed experimental protocols documented in Supplementary Note S4 and Fig S4. Test results are summarized in Figure 2E, where scatter points represent measured data and continuous lines correspond to model predictions.

Experimental results show type A achieves a maximum contraction ratio of 40%, with peak output forces of 11.9 N, 22.2 N, 34.6 N, 45.4 N, and 56 N under 30 kPa, 60 kPa, 90 kPa, 120 kPa, and 150 kPa respectively. Type B delivers a maximum contraction ratio of 44%, with peak forces of 10.3 N, 21.7 N, 34.9 N, 47 N, and 56.1 N at identical pressure levels. Type C also reaches a 44% maximum contraction ratio, generating peak forces of 21.6 N, 44.1 N, 69.4 N, 91.4 N, and 110.6 N under the respective pressures. Model calculations predict a maximum contraction ratio of 40.9% for Type A, with predicted peak forces of 11.7 N, 23.4 N, 35.1 N, 46.8 N, and 58.5 N. For Type B, the model predicts a 41.3% maximum contraction ratio and peak forces of 11.7 N, 23.4 N, 35.1 N, 46.8 N, and 58.5 N. Type C model outputs a 39.2% maximum contraction ratio and peak forces of 23.9 N, 47.8 N, 71.6 N, 95.5 N, and 119.4 N. Strong agreement between model predictions and experimental measurements verifies the accuracy of the proposed analytical framework for characterizing DPAM mechanical behavior.

For each DPAM variant, output force exhibits an approximately linear correlation with supply pressure. Comparative analysis across prototypes reveals type A and type B deliver nearly identical output forces, while type C generates roughly twice the force output of A and B. This demonstrates that fabric width $B_1$ dominates output force magnitude, whereas length $L_1$ exerts negligible influence on peak force.

Regarding force evolution with contraction ratio, output force remains nearly constant within a 32% contraction range, attributed to the stable airbag volume change rate during the cylindrical deformation stage. In contrast, conventional PAMs experience rapid force decay immediately upon contraction initiation. This consistent force output across a wide stroke range constitutes a significant performance advantage of DPAMs over existing designs.

## 6.3 Dynamic Characterization and Performance Comparisons

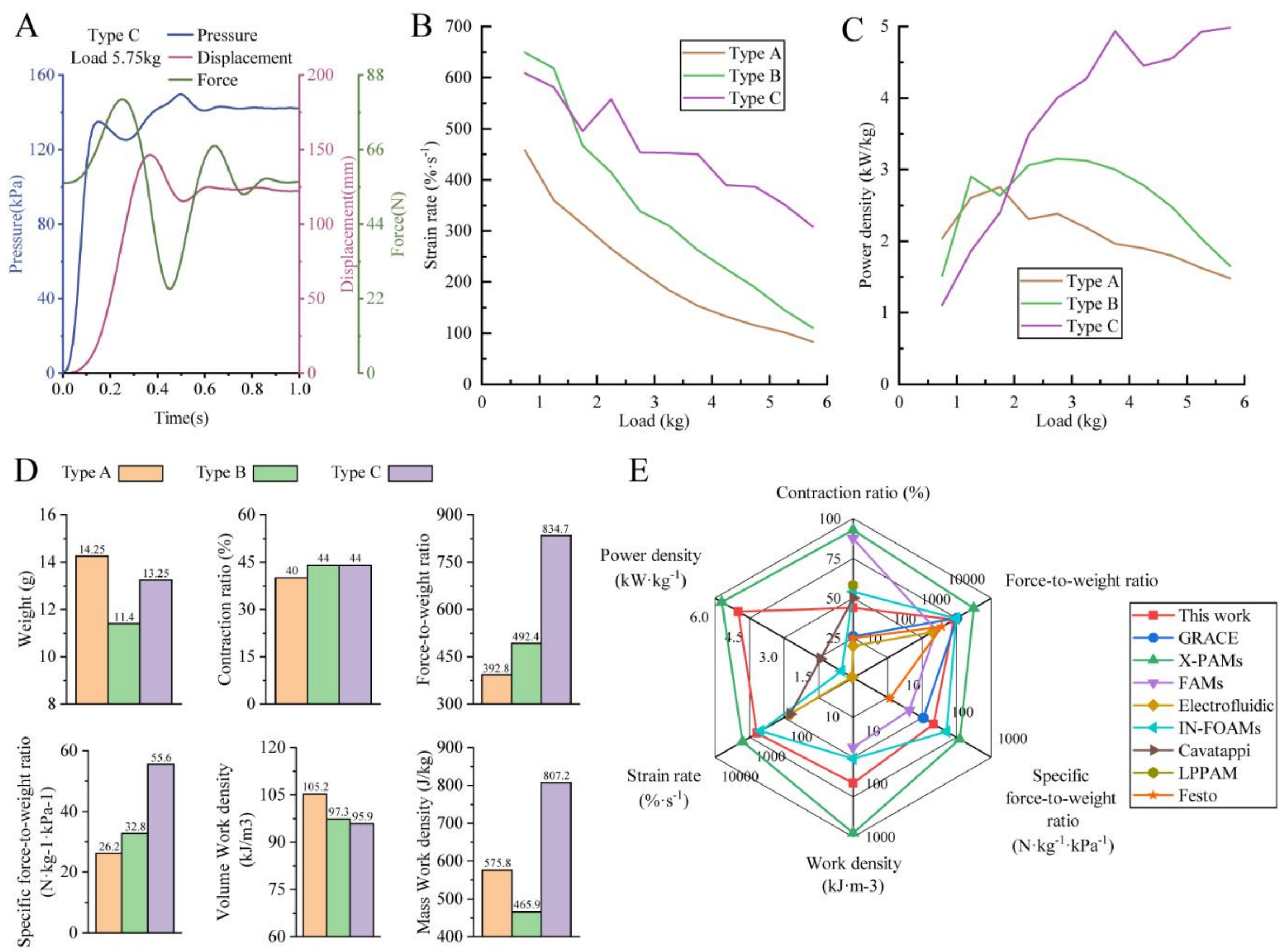

Fig. 3 DPAM dynamic testing and performance comparisons. (A) Dynamic output force and displacement response of the DPAM under step pressure input. (B) Strain rate under variable external loads. (C) Power density under varying loads. (D) Comparative performance metrics for DPAM variants, including mass, contraction ratio, force-to-weight ratio, specific force-to-weight ratio, volume work density, and mass work density. (E) Quantitative comparisons between DPAMs and state-of-the-art pneumatic artificial muscles across six core performance indicators.

The dynamic performance of DPAMs was characterized. Test hardware and protocols are detailed in Supplementary Note S5, Fig. S5, and Movie S3. Fig. 3A displays the transient response of type C lifting a 5.75 kg load under 140 kPa step pressure input. Initial displacement is zero, with static output force balancing the applied load. Motion initiates, reaching peak displacement of 149 mm at approximately 0.35 s, followed by mild oscillation and stabilization at 0.65 s. Output force peaks at 82.3 N at 0.25 s, undergoes two large oscillatory cycles, and stabilizes after 0.9 s. Full dynamic test datasets for type C under 0–5.25 kg loads and type A, B under 0–5.75 kg loads are provided in Fig. S6，Fig. S7, and Fig. S8.

Transient displacement and force data were post-processed to calculate strain rate and power density (Fig. 3B，Fig. 3C). Calculation methodologies are outlined in Supplementary Note S6. DPAM output power declines monotonically with increasing external load. Type A achieves a maximum strain rate of 457.7 $\%\cdot s^{-1}$ and a minimum of 83.3 $\%\cdot s^{-1}$. Type B exhibits a peak strain rate of 649.2 $\%\cdot s^{-1}$ and a minimum of 110.3 $\%\cdot s^{-1}$. Type C delivers a maximum strain rate of 608.8 $\%\cdot s^{-1}$ and a minimum of 308.7 $\%\cdot s^{-1}$. For type A and B, power density rises to a peak then decreases within the 0.75–5.75 kg load range: Type A reaches a maximum power density of 2.76 kW/kg, and type B peaks at 3.15 kW/kg. Type C's power density increases monotonically across the tested load range, attaining a maximum value of 4.98 kW/kg. Overall, type C demonstrates superior dynamic performance among the three variants, with type A exhibiting the weakest dynamic response. Therefore, wider, shorter airbag geometries are recommended to maximize DPAM actuation performance when mechanical constraints permit.

Fig. 3D tabulates key performance metrics including mass, contraction ratio, force-to-weight ratio, specific force-to-weight ratio, volume work density, and mass work density, with calculation formulas specified in Supplementary Note S6. Constructed from lightweight textile and plastic film substrates, DPAMs feature ultra-low mass: type A weighs 14.25 g, type B 11.4 g, and type C 13.25 g (mass measurements in Figure S9). All three prototypes achieve contraction ratios exceeding 40%. Type A has a force-to-weight ratio of 392.8, type B 492.4, and type C an exceptional 834.7, enabling lifting loads over 800 times the actuator self-weight without requiring ultra-high supply pressures (all tests conducted below 150 kPa). Specific force-to-weight ratios are 26.2 $N\cdot kg^{-1}\cdot kPa^{-1}$ for type A, 32.8 $N\cdot kg^{-1}\cdot kPa^{-1}$ for type B, and 55.6 $N\cdot kg^{-1}\cdot kPa^{-1}$ for type C. DPAMs deliver high energy density: volume work density converges to 100 $kJ/m^3$ across all variants, while mass work density varies drastically—type A reaches 575.8 J/kg, type B 465.9 J/kg, and type C 807.2 J/kg.

Figure 3E compares DPAMs against representative state-of-the-art PAMs across six critical evaluation criteria. DPAMs deliver mid-range contraction ratios, and they match leading PAMs in force-to-weight ratio, specific force-to-weight ratio, volume work density, strain rate, and mass power density. Full comparative parameter tables are listed in Supplementary Table S1.

## 6.4 DPAM Deployment for Bionic Robotic Applications

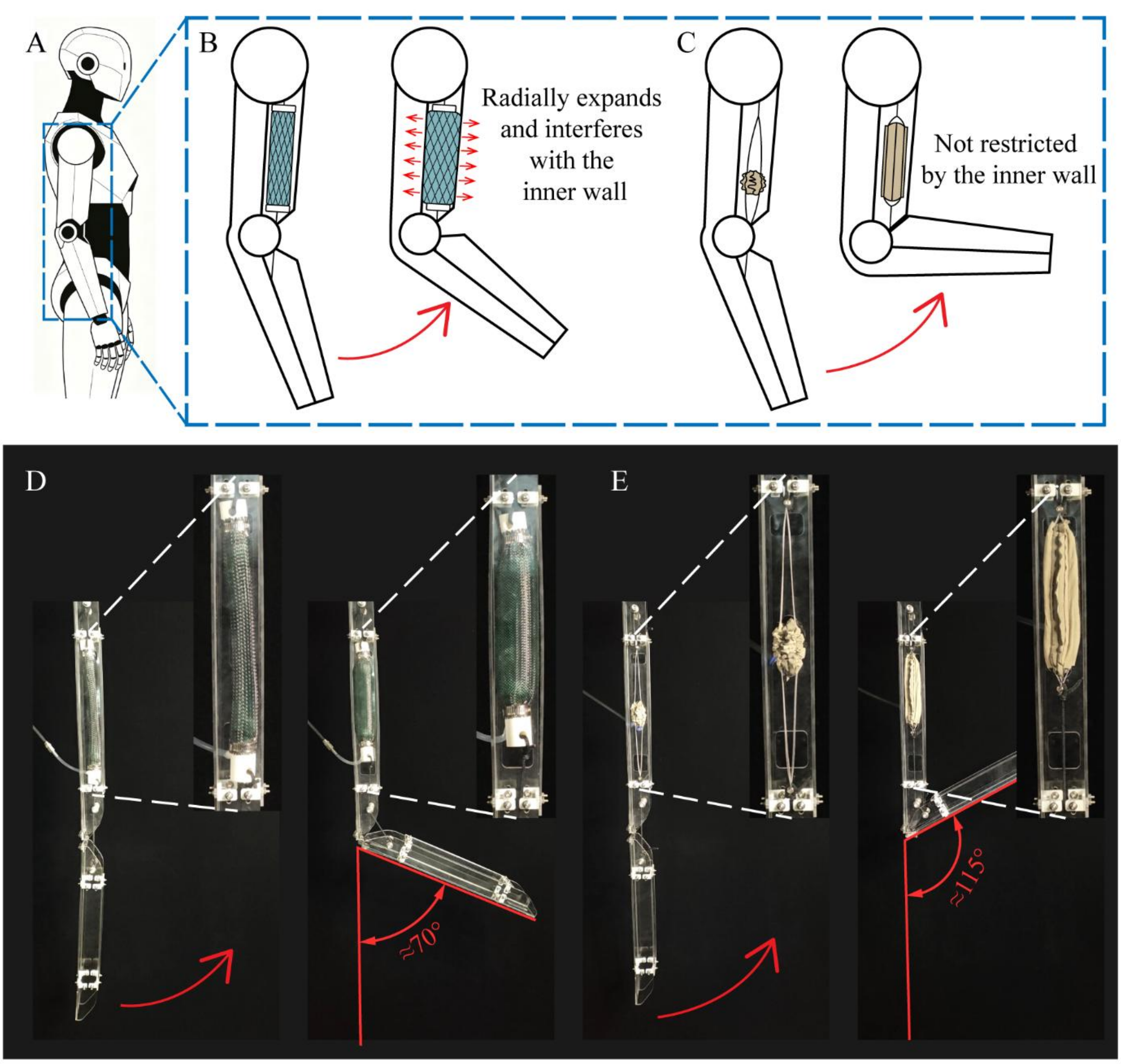

Fig. 4 DPAMs within bionic robotic systems. (A) Conceptual rendering of a DPAM-driven bionic robot. (B) Conventional PAMs generate radial expansion during contraction, colliding with the robot housing and limiting joint rotation range. (C) The DPAM eliminates radial expansion, enabling unconstrained full joint stroke without housing interference. (D) Transparent acrylic simplified bionic robot arm prototype. When driven by a McKibben PAM, radial expansion causes contact with the arm shell, restricting joint rotation to merely 70°. (E) The DPAM operates without radial interference, contracting fully to rotate the joint to 115°.

Pneumatic artificial muscles are widely adopted as actuation units for bionic robots. However, inherent radial deformation of conventional PAMs mandates reserved clearance to prevent mechanical interference during operation, sacrificing overall system compactness. The DPAM addresses this critical limitation.

As illustrated in Fig. 4A，B, when a traditional PAM actuates a bionic robot joint, axial contraction is accompanied by radial expansion. Insufficient housing clearance induces contact between the actuator and surrounding structural components, blocking full contraction and limiting maximum joint rotation angle. Using the DPAM eliminates this issue (Figure 4C): the absence of parasitic radial expansion allows full contraction within the arm housing without contact collisions, enabling complete joint rotation.

A simplified transparent acrylic bionic robot arm prototype was fabricated for intuitive comparative validation, enabling direct visual observation of internal actuator deformation. A custom McKibben PAM was manufactured for side-by-side testing. Installed inside the upper arm segment (Fig. 4D), the McKibben PAM expands radially upon inflation, compressing against the acrylic housing. Structural confinement prevents full contraction, limiting joint rotation to approximately 70°. In contrast, the DPAM mounted in the identical housing generates no radial expansion and avoids contact with surrounding components (Fig. 4E). Unrestricted full contraction drives the joint to a maximum rotation angle of 115°, substantially outperforming the McKibben counterpart. Supplementary Movie S4 records the complete comparative demonstration. This application case verifies that DPAMs enable drastically improved compactness in bionic robotic platforms.

## 6.5 DPAM Deployment for Industrial Production Machinery

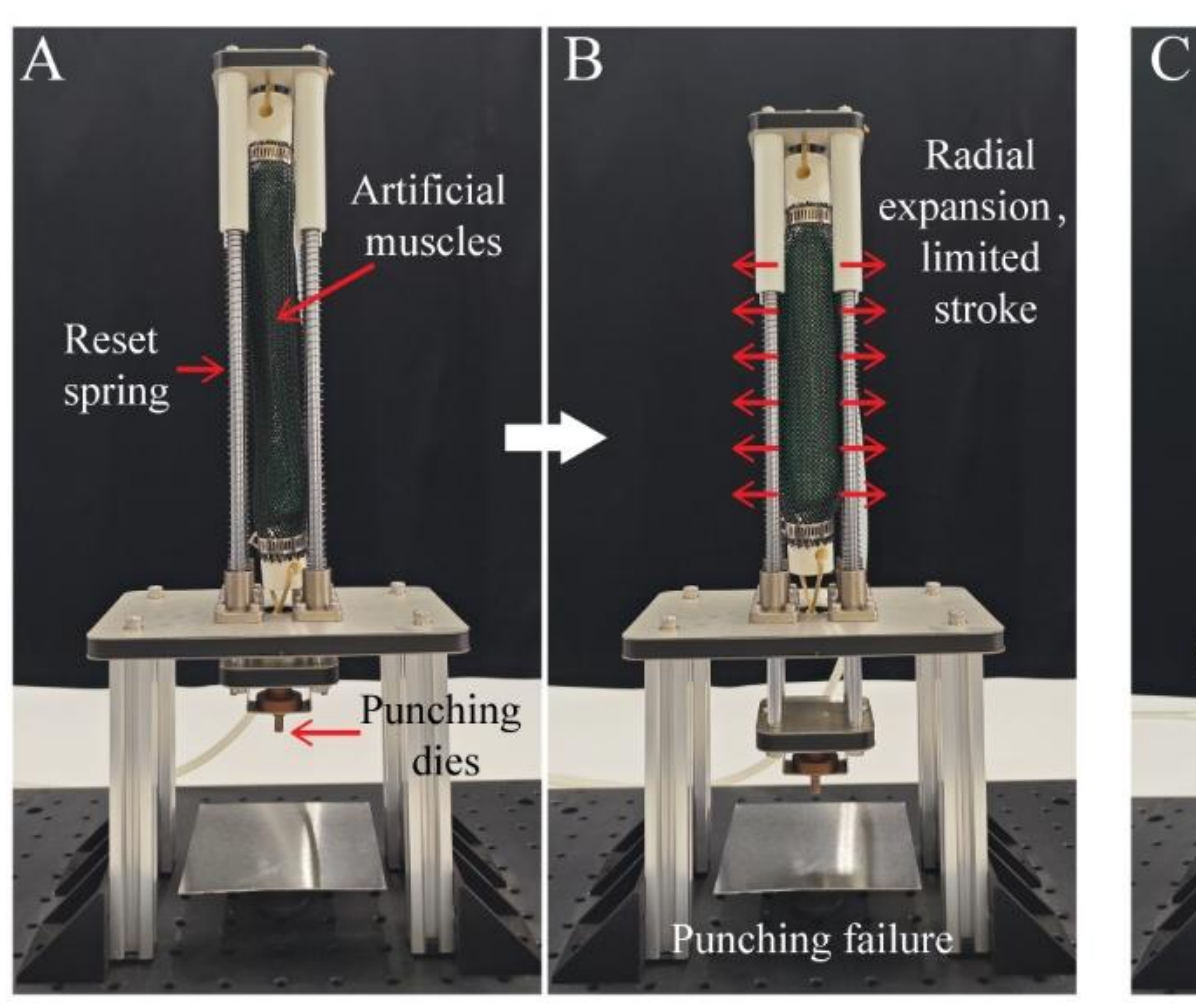

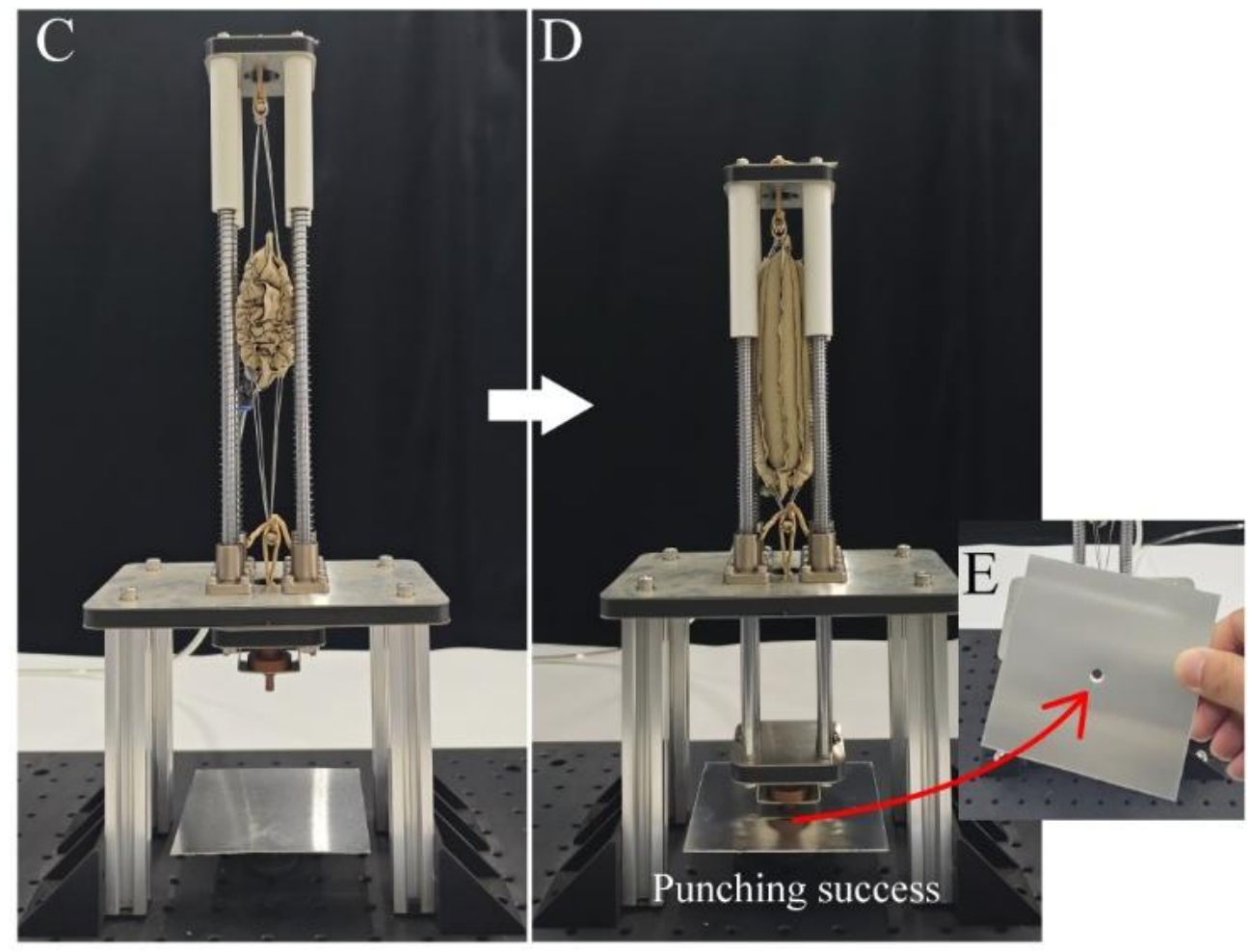

Fig. 5 DPAM integration in industrial production machinery. (A) Custom stamping machine prototype: PAM contraction drives the punch die downward to perforate metal sheets. Return springs reset the die upon depressurization. (B) Conventional McKibben PAMs generate radial expansion, interfering with guide shafts and return springs, limiting stroke and preventing successful punching. (C–D) The DPAM operates without radial deformation, eliminating component interference and delivering sufficient force and stroke to perforate aluminum sheets.

As lightweight alternatives to rigid pneumatic cylinders, PAMs see extensive deployment in industrial production lines. Compared with cylinders, PAMs feature lower mass and reduced material cost. Unlike bidirectional cylinders, PAMs only generate active contraction and rely on external elastic components for return motion. Conventional PAM systems require enlarged radial clearance to accommodate actuator expansion, reducing equipment compactness. DPAMs eliminate radial expansion entirely, removing the need for extra clearance and maximizing spatial efficiency.

A stamping machine was constructed for functional validation (Fig. 5A). A central PAM is surrounded by spring-loaded linear guide shafts, with a punching die mounted at the bottom. Pressurizing the PAM contracts the actuator, driving the die downward to punch aluminum sheets. Depressurization relaxes the PAM, and return springs retract the die to its initial home position.

Testing with a McKibben PAM (Fig. 5B) revealed severe radial expansion upon inflation, inducing collision between the actuator and surrounding guide shafts. Full contraction was blocked, resulting in insufficient die stroke that failed to perforate the aluminum sheet. When replaced with a DPAM (Fig. 5C， D), no radial expansion occured throughout the entire contraction cycle, eliminating mechanical interference with adjacent components. The die translated smoothly with sufficient kinetic energy to fully pierced the aluminum workpiece. Supplementary Movie S5 records the complete stamping demonstration. This industrial prototype validates that DPAMs significantly enhance the compactness of automated production equipment.

## 6.6 Scalability of DPAM

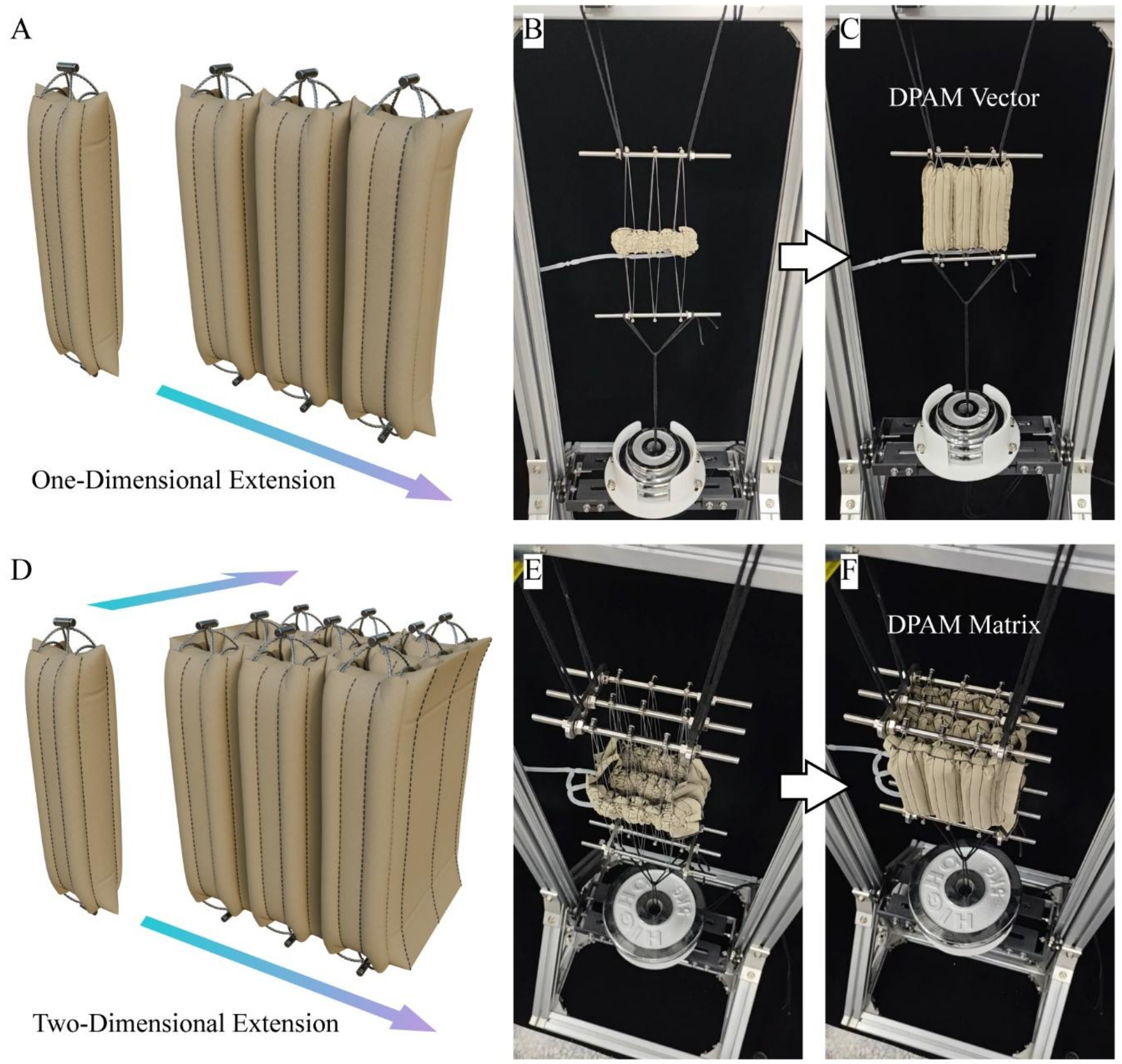

Fig. 6 Scalability of DPAM. (A–C) Single DPAM units can be stitched along their long edges to form one-dimensional actuator vectors of arbitrary count. A three-units DPAM vector was fabricated, delivering three times the output force of a single unit. (D–F) Multiple one-dimensional DPAM vectors were stitched along lateral edges with additional fabric substrates to form arbitrary two-dimensional actuator matrices. A 3×3 DPAM matrix was prototyped, generating nine times the force output of a single DPAM.

Scalability is a vital performance metric for actuators, determining adaptability across diverse application scenarios. Pneumatic cylinders exhibit excellent scalability, configurable as single-axis, multi-axis, or tabletop variants via minor mold modifications, enabling broad deployment across machinery. Conventional PAMs suffer poor scalability due to mutual radial interference when densely arranged in parallel arrays. The DPAM's zero-radial-expansion characteristic unlocks exceptional modular scalability.

As shown in Fig. 6A, a one-dimensional DPAM vector was fabricated via simple textile stitching techniques. Full manufacturing workflows are documented in Supplementary Note S7 and Fig S10. This work presents a three-units DPAM vector prototype, with vector length extendable to any positive integer. Physical hardware and load-lifting demonstrations are shown in Fig. 6B， C, with full motion recorded in Supplementary Movie S6. DPAM vectors resemble planar pouch motors but uniquely eliminate radial expansion, maintaining compact planar geometry while retaining >40% contraction ratios.

DPAMs support further two-dimensional assembly resulting in the DPAM matrix (Fig. 6D), with fabrication procedures detailed in Supplementary Note S8 and Figure S11. A square 3×3 DPAM matrix is demonstrated, with matrix dimensions customizable to arbitrary equal or unequal row/column counts. Triangular, hexagonal, and other irregular matrix layouts are also feasible. This geometric flexibility delivers design freedom comparable to pneumatic cylinders, accommodating complex spatial constraints across diverse applications. The fabricated 3×3 DPAM matrix (Fig. 6E, F) lifts a 10 kg load under merely 30 kPa supply pressure without radial expansion or collision between adjacent airbag units, achieving smooth, compact actuation. Supplementary Movie S6 records full matrix operation. Collectively, DPAMs demonstrate outstanding scalability for large-scale parallel actuation systems.

# 7 DISCUSSION

This work proposes a novel drawstring fabric pneumatic artificial muscle (DPAM). The DPAM adopts a three-component architecture: an inner airtight plastic film airbag, an outer load-bearing fabric shell, and plastic-coated steel drawstrings routed through fabric channels to constrict the airbag. All constituent materials are commercially available low-cost substrates. Despite its simple construction, the DPAM achieves the transformative advantage of zero parasitic radial expansion during axial contraction, drastically boosting system compactness from fundamental design principles. Meanwhile, the DPAM matches state-of-the-art PAM performance across all key metrics: 44% maximum contraction ratio, 834.7 force-to-weight ratio, 55.6 $N \cdot kg^{-1} \cdot kPa^{-1}$ specific force-to-weight, 95.9 $kJ/m^3$ volume work density, 807.2 J/kg mass work density, 608.8 $\% \cdot s^{-1}$ peak strain rate, and 4.98 kW/kg maximum power density. An analytical mathematical model describing the full two-stage deformation cycle is derived to accurately predict output force under varying geometric dimensions and operating pressures, providing systematic guidance for DPAM design optimization. Comprehensive static and dynamic testing across three distinct DPAM variants validate strong consistency between model predictions and experimental measurements.

Centered on its core advantage of eliminated radial expansion, two representative application cases—bionic robot arm and industrial stamping machinery—are presented to showcase practical advantages over conventional PAMs. The absence of radial expansion removes the requirement for reserved clearance between the DPAM and surrounding structural components, eliminating stroke-limiting mechanical interference. For bionic robots, DPAMs enable large joint rotation angles without enlarging arm housing diameter. For industrial stamping equipment, DPAMs deliver high-speed, long-stroke punch motion within radially compact enclosures comparable to rigid pneumatic cylinders, enabling reliable metal sheet perforation.

The modular scalability of DPAMs is further investigated. Free from radial collision risks, DPAMs are highly suited for dense parallel array deployment. One-dimensional DPAM vectors and two-dimensional DPAM matrices are developed via straightforward textile stitching processes with high geometric configurability. Modular DPAM arrays adapt to arbitrary workspace dimensions, broadening viable deployment scenarios.

The primary limitation of the current DPAM design lies in its minimum achievable size: miniature DPAMs cannot be fabricated using standard sewing techniques, restricting deployment in ultra-compact micro-scale applications. Future work will optimize manufacturing workflows and explore alternative non-textile substrates to overcome this miniaturization constraint.

Originating from the concept of textile-inspired soft actuator design—analogous to biological bionics—this study extracts mechanical functionality from the ubiquitous drawstring mechanism and successfully translates it into high-performance pneumatic artificial muscle hardware. The textile industry has undergone many years of development, which has accumulated a great deal of experience and wisdom. We anticipate that additional textile-derived mechanisms will inspire innovative solutions to unresolved challenges in soft robotics.

# 8 MATERIALS AND METHODS

## 8.1 Design and Fabrication of DPAMs, DPAM Vectors, and DPAM Matrices

SolidWorks software are used to generate dimensional drawings for plastic airbags and fabric airbags, including annotated thermal sealing zones and stitching paths for streamlined manufacturing. Plastic airbag designs integrate air tube ports with specified air-sealing schemes. Fabric stitching layouts reserve tubing passages and distribute rope channels uniformly. A thermal laminator (SJ2003, Youyalan, China) seals plastic film airbags. Air tube joints are sealed via sewing threads and reinforced with cyanoacrylate adhesive (402，loctite，China). A industrial sewing machine (8370，JUKI，Japan) stitches fabric shells and rope channels. 1 mm diameter plastic-coated steel wire serves as drawings, with rope terminals fixed using threaded metal sleeves. Full design and fabrication protocols are provided in supplementary materials.

## 8.2 DPAM Mathematical Modeling

The analytical model is derived from the principle of virtual work: pneumatic expansion work performed on the airbag equals mechanical output work delivered by the DPAM. Accurate calculation of DPAM output force relies on quantifying airbag volume variation throughout deformation. Due to the irregular inflated airbag geometry, geometric decomposition into a central cylindrical segment and two semicylindrical end caps is adopted for simplified volume integration. Continuity constraints are enforced at cylinder–semicylinder interfaces by adjusting semicylinder width. Complete mathematical derivation is documented in supplementary materials.

## 8.3 DPAM Experimental Characterization

An aluminum extrusion frame forms the universal test rig. A load cell (DYLF-102, DAYSENSOR, China) measures output force. A laser displacement sensor (HG-C1400-P, LGKG, China) captures actuator stroke. A pressure sensor (MPX5700GP, NPX, USA) monitors supply pressure. Sensor data are sampled via a multi-channel data acquisition card (USB3131A，ART, China). A precision pressure regulator (IR2010-02-A, SMC, Japan) adjusts input pressure, and an air compressor (ED-0204, Eidolon, China) supplies compressed air. Detailed experimental procedures are provided in supplementary materials.

## 8.4 Fabrication of the Bionic Robot Arm and Stamping Machine

Transparent acrylic sheets were laser-cut for the bionic robot arm prototype, assembled via

3D-printed connecting fixtures and bolt fasteners. Metal hinge joints serve as rotational axes, with ropes transmitting DPAM tensile force to drive joint rotation.

A custom McKibben PAM was fabricated for comparative testing, utilizing identical plastic film airbag material as the DPAM. Black nylon braided mesh forms the outer strain-limiting layer, paired with 3D-printed end fittings integrating barbed air hose connectors and rope routing holes. Hose clamps secure the airbag, braided mesh, and end fittings.

The stamping machine frame was assembled from aluminum extrusions, thin steel plates, and plastic panels, equipped with commercial punching dies (KOMAX China). Linear guide shafts and bearings constrain vertical die translation, with return springs providing reset force preloaded via 3D-printed adjustment sleeves.